\documentclass[10pt,twocolumn,letterpaper]{article}

\usepackage{cvpr}              

\usepackage{graphicx}
\usepackage{amsmath}
\usepackage{amssymb}
\usepackage{booktabs}
\usepackage{multirow}
\usepackage{makecell}

\usepackage[pagebackref,breaklinks,colorlinks]{hyperref}
\hypersetup{hypertex=true,
colorlinks=true,
linkcolor=red}

\def\cvprPaperID{} 
\def\confName{CVPR}
\def\confYear{2022}

\begin{document}

\title{Efficient Ray Sampling for Radiance Fields Reconstruction}

\author{Shilei Sun\\Ming Liu\\
Beijing Institute of \\
Technology\\
China\\
slsun@bit.edu.cn\\
bit411liu@bit.edu.cn
\and
Zhongyi Fan\\
Yuxue Liu\\
Chengwei Lv\\
Beijing Institute of Technology, China\\
zyfanzy@foxmail.com\\
18800197258@163.com\\
1612484269@qq.com
\and
Liquan Dong\\
Lingqin Kong\\
Beijing Institute of \\
Technology\\
China\\
kylind@bit.edu.cn\\
konglingqin3025@bit.edu.cn
}

\twocolumn[{
\renewcommand\twocolumn[1][]{#1}
\maketitle
\begin{center}
    \captionsetup{type=figure}
    \includegraphics[width=1\textwidth]{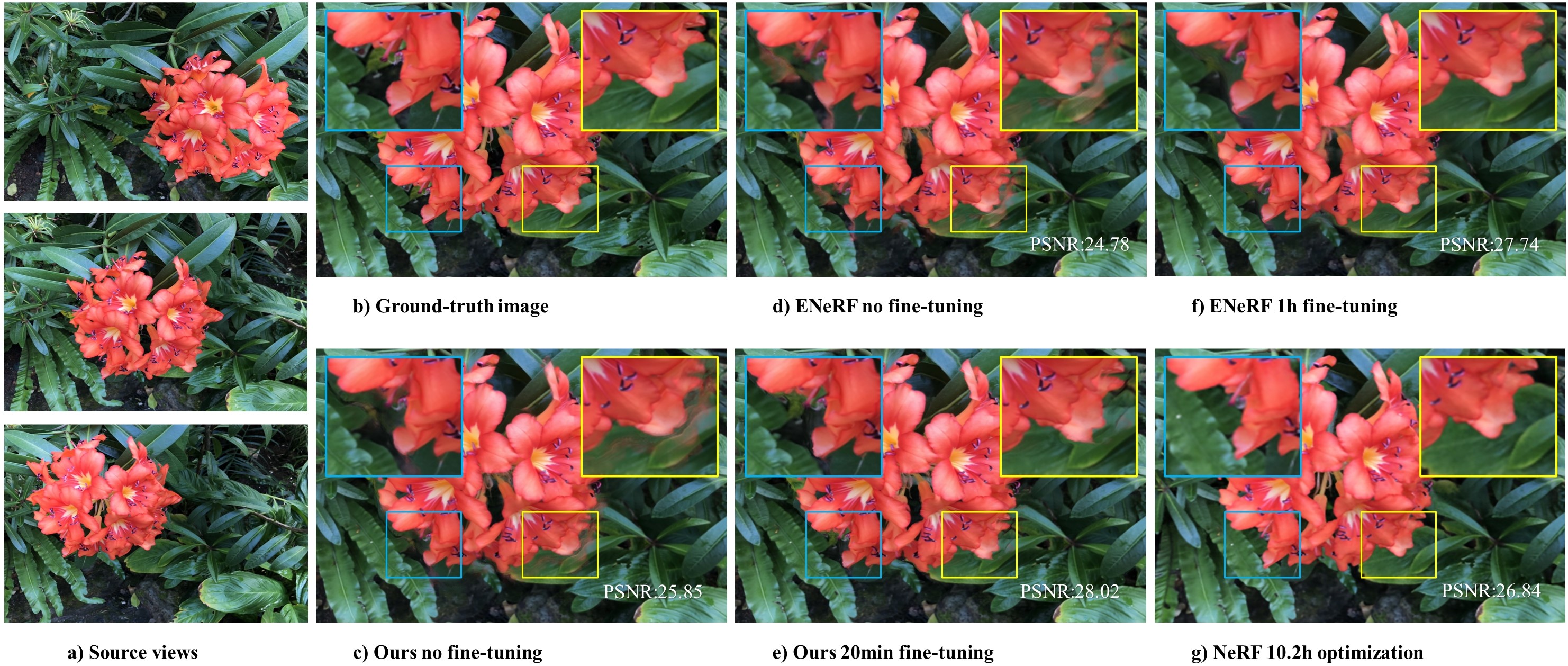}
    \captionof{figure}{\textbf{We incorporate a novel ray sampling method guided by pixel regions and depth into the existing ENeRF framework} \cite{25}. Leveraging the DTU dataset, we pretrain our model to generalize across varied scenes, enabling reconstruction of neural radiance fields from merely 3 input views (a), while reserving other views for testing (b). Our model synthesizes photorealistic novel views (c) qualitatively exceeding ENeRF (d). Although artifacts persist, it can be greatly improved by fine-tuning our reconstruction on three images for 20 min (e), which has better quality than the ENeRF's result from 1 hour fine-tuning (f) and NeRF’s result from 10.2h per-scen optimization (g).}
    \label{1}
\end{center}
}]

\begin{abstract}
 Accelerating neural radiance fields training is of substantial practical value, as the ray sampling strategy profoundly impacts network convergence.  More efficient ray sampling can thus directly enhance existing NeRF models' training efficiency. We therefore propose a novel ray sampling approach for neural radiance fields that improves training efficiency while retaining photorealistic rendering results. First, we analyze the relationship between the pixel loss distribution of sampled rays and rendering quality.  This reveals redundancy in the original NeRF's uniform ray sampling. Guided by this finding, we develop a sampling method leveraging pixel regions and depth boundaries. Our main idea is to sample fewer rays in training views, yet with each ray more informative for scene fitting. Sampling probability increases in pixel areas exhibiting significant color and depth variation, greatly reducing wasteful rays from other regions without sacrificing precision. Through this method, not only can the convergence of the network be accelerated, but the spatial geometry of a scene can also be perceived more accurately. Rendering outputs are enhanced, especially for texture-complex regions. Experiments demonstrate that our method significantly outperforms state-of-the-art techniques on public benchmark datasets.
\end{abstract}


\section{Introduction}
\label{sec:intro}

Neural radiance fields (NeRF) \cite{4} represent 3D scenes as implicit radiance fields fitted by neural networks. Given an arbitrary 3D location and viewing direction, NeRF outputs an emitted color and volume density. The training process of NeRF first samples pixels from source view images. Numerous spatial points are then sampled along each corresponding ray path. MLPs estimate volume density and color for each directional point. Finally, volume rendering integrates these outputs to predict pixel colors, which are compared against ground truth pixels to supervise network training.

However, a significant limitation of NeRF is the extensive time required to fit scenes. Initial NeRF training on a novel scene with a single GPU spanned approximately two days. Subsequent works \cite{22,61,62} revealed the bottleneck is not the learning radiance field itself, but inefficient oversampling of empty space during training. This aligns with the high sparsity of real 3D scenes. Yet NeRF's initialization assumes a uniform density distribution, mismatched to the distributions of real-world surface.From this perspective, we expect that by learning the density function of NeRF to simulate the surface distribution in the real world, the fitting efficiency of NeRF to the scene geometry can be further improved.

Analysis of multiple NeRF models' loss distributions during training reveals slower fitting efficiency in image regions with drastic color variations compared to other areas. This discrepancy manifests in 3D scene regions exhibiting significant detail texture and depth variations. Therefore, we propose the redundancy hypothesis for ray sampling in neural radiance fields. In addition to oversampling empty space, we find suboptimality in the ray sampling scheme itself during training. As 3D space is incrementally fitted, radiance fields first estimate the flat and low-texture areas more accurately, before less efficient fitting of regions with substantial depth and color changes. This follows the intrinsic fitting sequence of deep neural networks, which encode low-frequency information first, followed by the high-frequency information contained in intricately detailed textures.

Based on this hypothesis, we introduce a ray sampling strategy guided by pixel regions and depth boundaries. The key idea is to train with fewer sampled rays, yet more informative for scene fitting within source views. We increase sampling probability in areas with significant color and depth variations, greatly reducing redundant rays from other pixels to accelerate network fitting and improve efficiency. Our method enables faster per-scene optimization, achieving the same convergence effect in minutes through fine-tuning reconstructions from given views. Moreover, our sampling strategy better perceives spatial geometry, improving rendering quality especially for rich-texture regions. Since ray sampling is ubiquitous in NeRF, our strategy can directly transfer to existing models. We demonstrate this by applying our approach to ENeRF \cite{25}, achieving state-of-the-art performance on both view synthesis and training time, as shown in Fig. \ref{1}.


\section{Related work}
\label{sec:work}
\noindent
\textbf{Novel view synthesis.} View synthesis methods typically rely on intermediate 3D scene representations. NeRF \cite{4} pioneered photorealistic novel view synthesis with implicit scene functions. Unlike previous methods, NeRF attempts to recreate an implicit volume as the intermediate representation. Improvements based on the problems of NeRF include accelerating the training \cite{27,28,29,30,31,32} and rendering \cite{22,23,24,25,26} process, targeting on dynamic scenes \cite{33,34,35,36}, better generalization \cite{37,38,39,40,25}, training with fewer viewpoints \cite{41,42,36}. There are also some researches on the application of NeRF. The process of estimating different model parameters (camera, geometry, materials, lighting parameters) from realistic data is called inverse rendering, which purpose is to estimate the pose of camera \cite{43,44}, edit materials and lighting \cite{45}. Controlled editing is a top priority in computer vision, and the development of NeRF in this area can be described as a separate branch, from EidtNeRF \cite{46} to GIRAFFE \cite{47} (CVPR2021 Best Paper). These methods are mainly combined with GAN \cite{48} and NeRF to achieve controlled editing, and the combination of diffusion model \cite{49} and NeRF is becoming increasingly popular \cite{50,51,52}. Another important area is digital human body, which includes the Facial Avatar \cite{53,54} and Human Bodies \cite{34,36,25}. After a long time of development, NeRF has shown better and better performance for view synthesis.

\begin{figure*}
    \centering
    \includegraphics[width=0.9\linewidth]{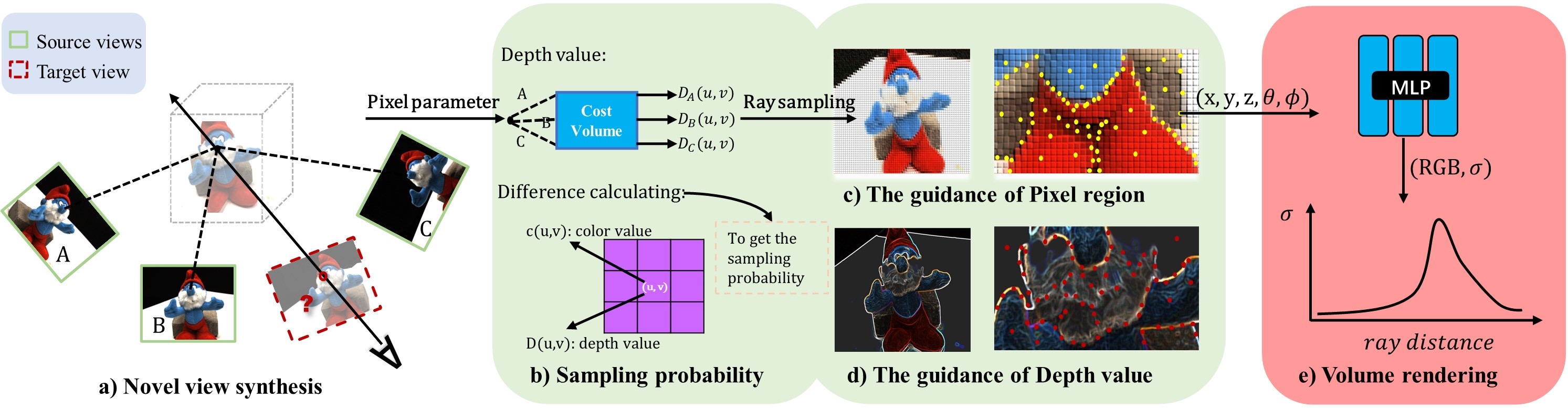}
    \caption{ \textbf{Overview of our ray sampling model.} To render a novel target view, multi-view source images of static or dynamic scenes at one frame are utilized (a). The cost volume \cite{25} is leveraged to obtain depth information, and we use it and color value to get the sampling probability of the pixel $(u,v)$ (b). With the sampling probability of pixels from source view images $I_i$, during ray sampling, our model samples additional rays in pixel areas exhibiting substantial color variation (c). Similarly, guided by the depth-based sampling strategy, more rays are sampled in regions where geometric depth varies markedly (d). Finally, through volume rendering (e), our network outputs colors and densities along the sampled rays to render pixel color.}
    \label{3}
\end{figure*}

\noindent
\textbf{Training and rendering acceleration.} NeRF's substantial computational requirements for training and rendering motivate research into efficiency improvements. For rendering acceleration, octrees-based \cite{22} approaches decompose NeRF into numerous small MLPs to enable model distillation. Some methods \cite{55,56,25} improve the sampling strategy to accelerate the rendering process. ENeRF \cite{25} uses geometric feature bodies to guide the radiance fields sampling near surfaces, significantly improving the training and rendering efficiency. For training acceleration, DVGO \cite{59} employs explicit discrete volume representations to rapidly simulate radiance fields, reducing training time to a few minutes. Plenoxels \cite{30} uses a sparse voxel grid and spherical harmonics for fast viewpoint rendering. Instant-NGP \cite{29} applies a multiresolution hash encoding and optimized GPU implementation to train high-quality neural graphics primitives in seconds. In summary, the methods of rendering and training acceleration make NeRF more practical, from optimized model architectures and sampling strategies to discrete representations and GPU-optimized implementations. 

However, current acceleration methods predominately focus on static scenes, with limited work on modeling dynamics \cite{25,60,61}. Recently, Fourier PlenOctrees \cite{61} extends PlenOctrees \cite{22} to dynamic settings by representing time-varying density and color in the frequency domain. However, this approach necessitates intensive view sampling and prolonged training. In contrast, our proposed method enables efficient training for dynamic scene using only sparse camera viewpoints.

\noindent
\textbf{Ray sampling methods.} The conventional ray sampling strategy of neural radiance fields utilizes uniform distribution \cite{4}, with evenly distributed ray locations across training images. Later work introduced adaptive sampling \cite{42,25,54} guided by rendering loss to focus on inaccurately rendered regions. This approach is analogous to sample mining in deep learning training, where pixels exhibiting poor convergence are identified via loss map to receive additional sampling focus. There are also some ray sampling methods that sample fewer points in per-ray. \cite{57,58,39} achieve acceleration by predicting surfaces and sampling near the surfaces, reducing the number of necessary sampling points for rendering at each ray. Uniquely, we sample varying ray densities based on depth boundaries and pixel regions. This simultaneously accelerates network convergence and improves geometry perception.

\section{Method}
\label{sec:method}

Depth information of rough scenes is obtained from the depth estimation module in ENeRF \cite{25}. An overview of our proposed method is illustrated in Fig. \ref{3}. First, the rendering process of neural radiance fields is introduced (Sec. \ref{3.1}). Then, through analysis of observations from several experiments in the training stage of neural radiance fields, we propose a redundant hypothesis regarding ray sampling for neural radiance fields (Sec. \ref{3.2}). By combining estimated scene depth information and image pixel distribution, we devise a ray sampling strategy guided by pixel regions (Sec. \ref{3.3}) and depth boundaries (Sec. \ref{3.4}). The entire network is trained in an end-to-end manner using RGB images as supervision (Sec. \ref{3.5}).

\subsection{Preliminaries}\label{3.1}

Description of the entire framework of neural radiance fields (NeRF) : Given a 3D location \!$p=(x,y,z) \!\in\! \mathbb{R}^3$\! and its associated 2D viewing direction \!$d=(\theta,\phi) \!\in\!  \mathbb{R}^2$\!, the MLP network \!$F_\Theta\!:\!(p, d)\!\rightarrow\!(c, \sigma)$\! maps the 5D coordinate point $(p, d)$ to its corresponding volume density $\sigma$ and directional emitted color $c=(r,g,b)$.  The network parameters are then optimized by minimizing the color error at the pixel level from volume rendering.

During the training stage, a certain number of 5D coordinate points are sampled from each ray $r_i^k\in \mathrm{R}_i$, where $r_i^k$ represents the $kth$ ray taken from the $ith$ training image $I_i$, and $\mathrm{R}_i$ is the set of rays generated from images $I_i\in I,i=1,2,....,N$ (where $N$ is the number of trained images). The volume rendering $V_\Theta:r_i^k\rightarrow \hat{c}_i^k$ then makes color predictions at the pixel level along the colors and densities of the sampled ray to obtain the color value of each pixel $\hat{c}_i^k$. The final optimization goal of the model is to minimize the volume rendering loss from all the sampled rays, as follows:
\begin{equation}
    min_\Theta \mathcal{L}=\frac{1}{NM} \sum_{r_i^k\in \mathrm{R}_i}||V_\Theta(r_i^k)-c_i^k||_2^2,
\end{equation}
where $c_i^k$ is the true color of pixels, $M$ is the number of $\mathrm{R}_i$, which represents the number of sampled rays per image.


\subsection{Efficiency of ray sampling}\label{3.2}

This section describes the problem of ray sampling in volume rendering from a holistic perspective. The key question is how to obtain all the sampled rays for the training images. Generally, the method based on classical radiance fields randomly selects $M$ pixels in image $I_i$ and emits $M$ rays from the selected pixels along the observed direction. Since the ray parameters depend solely on the pixel position and viewing direction, $r_i(u,v)$ can represent a ray originating from pixel $(u,v)$. The ray direction aligns with the viewing direction of image $I_i$. In this manner, the positions of the sampled rays follow a uniform distribution across training images:
\begin{equation}
    r_i(u,v) \sim  U(I_i),u\in[0,H_i],v\in[0,W_i],
\end{equation}
where $H_i$ and $W_i$ are the height and width of image $I_i$, respectively. However, this distribution fails to match the real-world surface, which is defined as $\alpha \sim E(w)$, where $\alpha$ is the scene density. $E(w)$ represents the density distribution that depends on scene geometry and appearance.

The uniform ray sampling strategy is effective in practice, but there remain some potential issues. First, this approach fails to fully capture contextual information in images, specifically the non-uniform distribution of color differences among pixels. In fact, there exist numerous minute regions in an image where successive pixels tend to exhibit similar colors. The color variation between certain pixel areas and their neighbors may be negligible, hence the sampled rays acquire less information in these regions.

Moreover, depth variation within a scene has a substantial impact on image information. Compared to regions with relatively flat depth, areas in a scene where depth undergoes abrupt changes contain more high-frequency signals. Rays in these regions necessitate sampling a greater number of 3D points to accurately determine their spatial depth. In trained radiance fields, this phenomenon commonly manifests as blurring of edges of three-dimensional objects. Consequently, in regions with flat color and depth variation, the rendering loss rapidly converges as the density of most sampled points along the rays approaches zero. In such regions, sampling fewer rays suffices to achieve a more precise fit of the radiance fields. In contrast, regions with stronger color and depth variation encompass more intricate scene information, necessitating sampling additional rays and spatial points to capture fine details.

Therefore, during the training phase, sampling a substantial number of rays is unwarranted for regions with relatively simple structure. Instead, focus should be directed toward areas with intricate structure. Loss in regions of intricate structure is always optimized more slowly. This demonstrates that existing methods of ray sampling during NeRF training contain redundancies, directly impacting the convergence efficiency of radiance fields.

Based on the preceding analysis, two sampling strategies are proposed to optimize the distribution of sampled rays for input images. The first strategy calculates the prior probability distribution based on the color of pixel region, while the second strategy dynamically adjusts the ray sampling distribution utilizing estimated depth. And we can make a learning density function of NeRF to simulate the surface distribution in the real world
\begin{equation}
    r_i(u,v) \sim P_{samp}(I_i)=P_c(I_i)+P_d(I_i),
\end{equation}
where $P_{samp}$ is the probability density function, which can guide the radiance fields to sample rays. $P_c$ and $P_d$ represent the the probability density functions of pixel-guided and depth-guided, respectively.

\subsection{Sampling ray with the guidance of pixel region}\label{3.3}

The information density across different regions of an input-view image largely depends on the intensity of color and depth variation. To identify high and low information density areas in the image, color variation in pixel regions is leveraged to capture image information and guide non-uniform ray sampling in radiance fields. Intuitively, a pixel exhibiting identical color to its neighborhood likely contains less information, whereas a pixel with substantial color differences from its surroundings encompasses more scene details. Therefore, we quantify the information richness level by computing variation intensity within pixel neighborhoods, subsequently replacing the uniform distribution of sampled rays during training.

In the ray sampling stage, we design a probability density function $P_c(u,v)$ that maps the position of pixel $(u,v)$ to a prior probability, denoting the likelihood of sampled rays at that location. We define $P_c$ as the standard deviation of a $n \times n$ neighborhood of pixel colors $c$ centered on pixel $(u,v)$, as follows:
\begin{equation}
\begin{aligned}\label{e3}
 P_c(u,v)=std(c(u,v))=\sqrt{\frac{1}{n^2}\sum_{x^{\prime},y^{\prime}}[c(x^{\prime},y^{\prime})-\Bar{c}]^2},\\
    x^{\prime} \in {[u-\frac{n-1}{2}, u+\frac{n-1}{2}]}, \\
    y^{\prime} \in {[v-\frac{n-1}{2}, v+\frac{n-1}{2}]},     
\end{aligned}
\end{equation}
where $\Bar{c}$ denotes the mean color of the $n \times n$ pixels centered on pixel $(u,v)$. In the experiments, we set $n=3$.

To facilitate computation, the standard deviation equation is utilized for code implementation:
\begin{equation}
   std(c(u,v))=\sqrt{E(c(x^{\prime},y^{\prime})^2)-E(c(x^{\prime},y^{\prime}))^2}.
\end{equation}

If pixel $(u,v)$ exhibits identical color to its neighbors, then $P_c(u,v)=0$. Conversely, if the pixel color differs substantially from the surrounding color, $P_c(u,v)$ will be higher, denoting this position contains more scene details. In the 2D pixel space, pixels with a higher value of $P_c$ typically corresponds to this area with sharper color variation. Correspondingly, in 3D space, some of these pixels match planes with richer texture details, while others match locations with sharper density variation, representing the boundary surfaces of 3D objects. To balance the discrepancy between maximum and minimum values for $P_c$, normalization is necessitated, as depicted:
\begin{equation}
    P_c^{\prime}(u,v)=\frac{clamp(s,max(P_c(u,v)))}{max(P_c(u,v))},
\end{equation}
where the threshold $s$ is defined as $0.01 \times mean(P_c(u,v))$. Values below $s$ will be scaled to $s$ to avoid sampling too few rays at certain locations, which can prevent model underfitting. The normalization function $P_c^{\prime}$ generates a specific probability distribution for each input source view. After normalization, distribution $P_c^{\prime}(u,v)$ is within the interval $[0,1]$, then the distribution $P_c^{\prime}$ can substitute the uniform distribution to sample rays utilized for NeRF training:
\begin{equation}
     r_i(u,v)\sim P_c^{\prime}(u,v),u\in [0,H_i],v\in [0,W_i],
\end{equation}
where $r_i(u,v)$ represents the sampled ray from pixel $(u,v)$ of image $I_i$. More details about $clamp$ are in the supplementary material.

\subsection{Depth-guided ray sampling}\label{3.4}

Analogous to the guided sampling of pixel regions delineated in Sec. \ref{3.3}, we devise a depth-guided ray sampling strategy based on the scene depth probability distribution. The depth prediction follows the coarse-to-fine approach in ENeRF \cite{25}. In ENeRF, for pixel $(u, v)$ situated in the target view's feature map, linear interpolation of the depth probability volume $P_i(u,v)$ is requisite to ascertain its probability at a specific depth plane $L_i$. The depth value $D(u,v)$ at pixel $(u,v)$ constitutes the expectation of the depth probability distribution, where $L_i$ denotes the different depth planes sampled from $[L_{min}, L_{max}]$:
\begin{equation}\label{desired_depth}
    D(u,v)=\sum_{i=1}^{N_d}{P_i(u,v)\cdot L_i(u,v)},
\end{equation}
where $N_d$ is the number of sampled depth planes.

Applying the scheme analogous to Eqn. \ref{e3} on the desired depth map of the rendering view (from Eqn. \ref{desired_depth}) obtains the sampling probability distribution guided by the depth boundary 
\begin{equation}
    P_d(u,v)=std(D(u,v))=\sqrt{\frac{1}{n^2}\sum_{x^{\prime},y^{\prime}}[D(x^{\prime},y^{\prime})-\Bar{D}]^2},
\end{equation}
where $\Bar{D}$ denotes the mean depth value. The $x^{\prime}$ and $y^{\prime}$ is also same as Eqn. \ref{e3}. The same normalization operation is utilized to avoid underfitting for certain rays, getting the normalization $P_d^{\prime}(u,v)$.

In the training stage, two sampling strategies guided by pixel region and depth boundary impact the ray sampling process through weighted fusion, as delineated in the following equation:
\begin{equation}
    P_{samp}(u,v)=\beta P_c^{\prime}(u,v)+(1-\beta)P_d^{\prime}(u,v),
\end{equation}
where $\beta$ represents a weight parameter that modulates the proportional contributions of the two probability distributions for fusion. Given that the depth information is progressively fitted during training, the value of $\beta$ incrementally rises from 0 to 0.5 across training iterations. Please refer to the supplementary material for further details on the analysis of sampling efficiency and methodology.


\subsection{Training}\label{3.5}

We integrate our ray sampling method, guided by pixel region and depth boundary, into ENeRF \cite{25}. During training, our ray sampling approach samples fewer rays per iteration than ENeRF, while achieving improved performance. Consequently, the ENeRF model employing our approach occupies less GPU memory. Thus, $\mathcal{L}_{mse}$ and $\mathcal{L}_{perc}$ can be utilized for supervision:
\begin{equation}
\begin{aligned}
 \mathcal{L}_{mse}=\frac{1}{N_r}\sum_{i=1}^{N_r}\parallel{\hat{C_i}-C_i}\mid\mid^2,\\
     \mathcal{L}_{perc}=\frac{1}{N_i}\sum_{i=1}^{N_i}\parallel\Phi(\hat{I_i})-\Phi(I_i)\parallel,    
\end{aligned}
\end{equation}
where $\mathcal{L}_{mse}$ represents the mean square error between the color of rendered pixels and ground-truth pixels. $\mathcal{L}_{perc}$ denotes perceptual losses obtained by sampling image blocks. $N_r$ signifies the number of sampled rays per iteration and $N_i$ denotes the number of image blocks. $\Phi$ represents the definition of perceptual function \cite{25}. $\hat{C}_i$ and $C_i$ signify the rendered and ground-truth colors, respectively.

The final loss function is as follows:
\begin{equation}
    \mathcal{L}=\mathcal{L}_{mse}+\lambda^{\prime}\mathcal{L}_{perc}.
\end{equation}
In the experiments, we set $\lambda^{\prime}=0.01$.

\begin{figure*}
    \centering
    \includegraphics[width=0.9\linewidth]{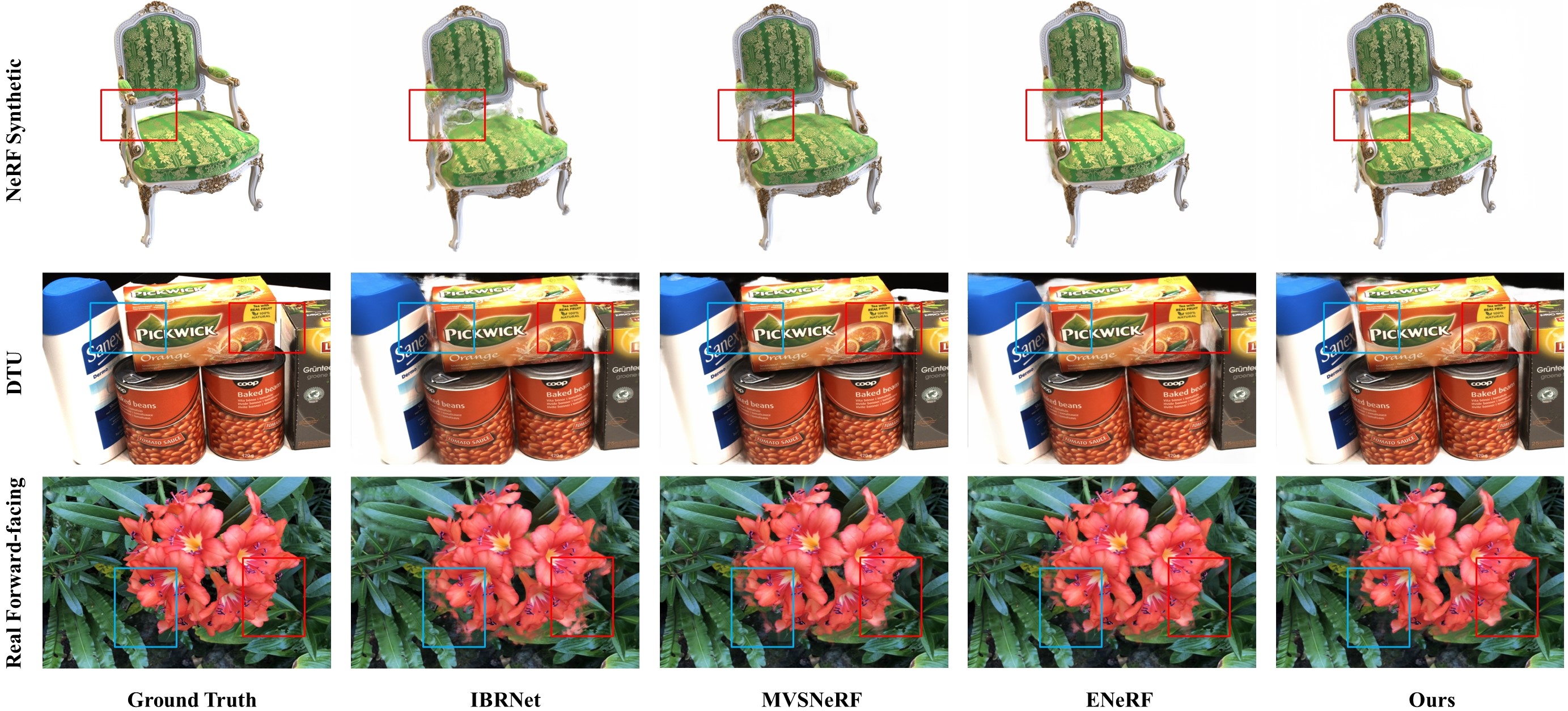}
    \caption{ \textbf{Qualitative comparison of view synthesis results under the generalization setting.}}
    \label{4}
\end{figure*}

\section{Experiments}
\label{sec:expe}

\begin{table*}[t]
    \renewcommand{\arraystretch}{1}
    \centering
    \caption{\textbf{Quantitative results of static scenes.} ``Generalization'' means that the model is not additionally fine-tuned. At the bottom, results are presented of different baselines fine-tuning their models on specific scenes. Our approach attains the optimal performance on PSNR and LPIPS, alongside outstanding results on SSIM. The baseline results are obtained from ENeRF \cite{25}.}
    \label{t1}
    \scalebox{0.85}{
    \begin{tabular}{c|c|ccc|ccc|ccc}
    \Xhline{1.2pt}
        \multirow{2}{*}{Methods} & \multirow{2}{*}{\makecell{Training settings}} & \multicolumn{3}{c|}{DTU} & \multicolumn{3}{c|}{NeRF Synthetic} & \multicolumn{3}{c}{Real Forward-facing} \\
        \cline{3-5} \cline{6-8} \cline{9-11}
        ~ & ~ & PSNR$\uparrow$ & SSIM$\uparrow$ & LPIPS$\downarrow$ & PSNR$\uparrow$ & SSIM$\uparrow$ & LPIPS$\downarrow$ & PSNR$\uparrow$ & SSIM$\uparrow$ & LPIPS$\downarrow$ \\ 
        \hline
        IBRNet & \multirow{4}{*}{\makecell{Generalization \\ (only pre-trained)}} &26.04&0.917&0.191&22.44&0.874&0.195&21.79&0.786&0.279 \\
        MVSNeRF & ~ &26.63&0.931&0.168&23.62&0.897&0.176&21.93&0.795&0.252 \\ 
        ENeRF & ~ &27.61&\textbf{0.956}&0.091&26.65&0.947&0.072&22.78&0.808&0.209 \\
        Ours & ~ &\textbf{27.85}&0.941&\textbf{0.059}&\textbf{26.82}&\textbf{0.949}&\textbf{0.062}&\textbf{23.01}&\textbf{0.821}&\textbf{0.194} \\
    \hline
        IBRNet(1h) & \multirow{6}{*}{\makecell{Fine-tuning \\ on given scene}} &\textbf{31.35}&0.956&0.131&25.62&0.939&0.111&24.88&0.861&0.189 \\
        MVSNeRF(15min) & ~ &28.51&0.933&0.179&27.07&0.931&0.168&25.45&0.877&0.192 \\ 
        ENeRF(15min) & ~ &28.73&0.956&0.093&27.20&0.951&0.066&24.59&0.857&0.173 \\
        ENeRF(1h) & ~ &28.87&\textbf{0.957}&0.090&27.57&\textbf{0.954}&0.063&24.89&0.865&0.159 \\
        Ours(15min) & ~ &29.30&0.953&0.092&27.32&0.943&0.066&25.04&0.861&0.166 \\
        Ours(30min) & ~ &30.61&0.956&\textbf{0.088}&\textbf{27.62}&0.952&\textbf{0.055}&\textbf{26.19}&\textbf{0.882}&\textbf{0.0145} \\
    \Xhline{1.2pt}
    \end{tabular}
    }
\end{table*}

\subsection{Implementation Details}

\begin{figure*}
    \centering
    \includegraphics[width=0.9\linewidth]{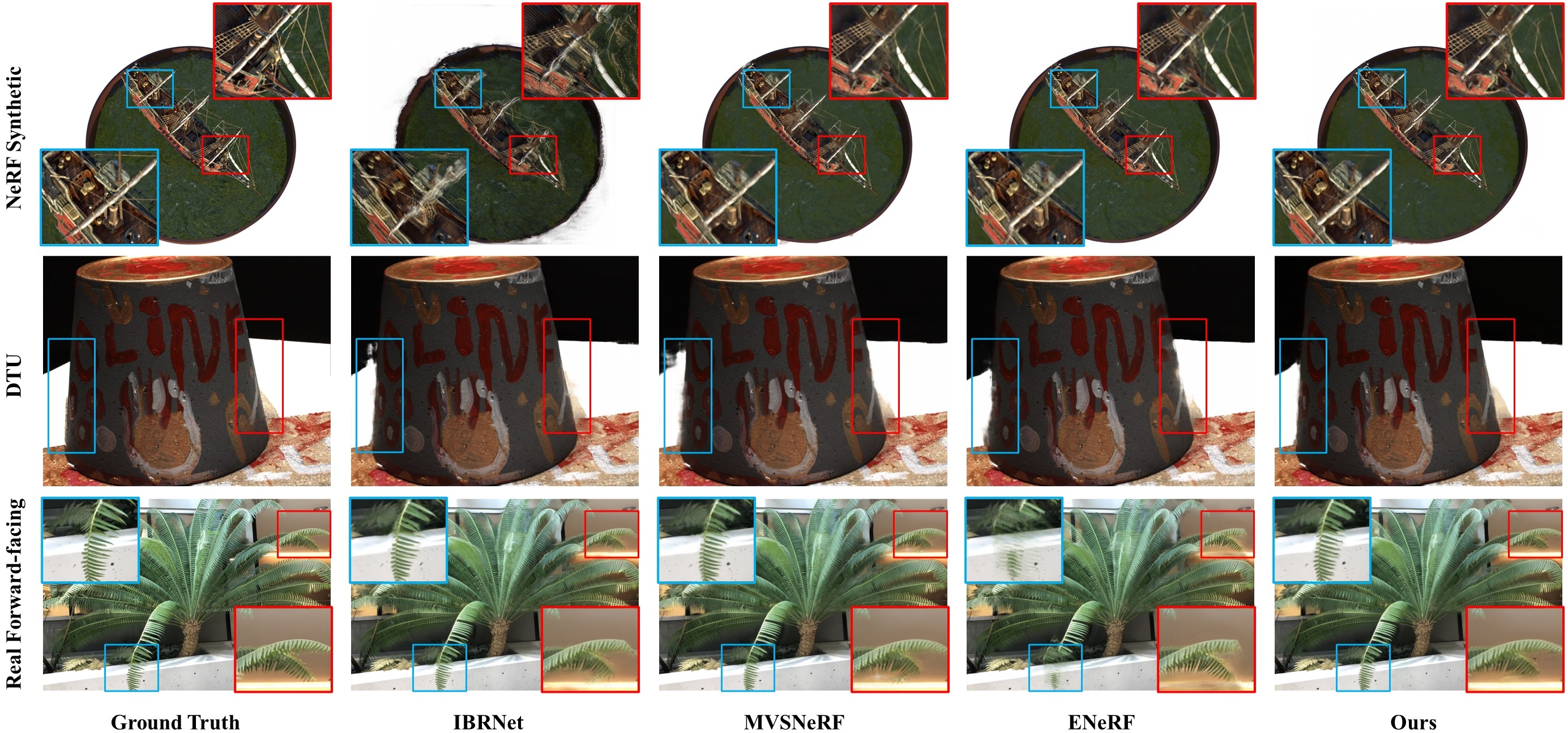}
    \caption{ \textbf{Qualitative comparison of view synthesis results under fine-tuning on the given scene setting. }}
    \label{5}
\end{figure*}

\noindent
\textbf{Experiments.} For the ENeRF model \cite{25} utilizing our ray sampling strategy, the same experimental setup is followed. The model is exclusively trained on the DTU dataset \cite{66} to accomplish rapid generalization for novel scenes. Subsequently, the model undergoes fine-tuning for a few epochs on specified novel scenes. Rays are sampled under the guidance of pixel region and depth boundary, where 2 spatial points are sampled along each ray. In experiments, ablation studies are conducted to assess the performance of various ray sampling strategies on rendering outcomes.

\noindent
\textbf{Datasets.} We use the same train-test split and evaluation setting from ENERF \cite{25}. For each test scene, 16 views are chosen as seen views and the remaining 4 views as novel views for evaluation. The framework is pre-trained on the DTU dataset to learn generalizable neural radiance fields. Real Forward-facing and NeRF Synthetic datasets \cite{4} are also tested. For dynamic scenes, the ZJU-MoCap \cite{34} and our produced digital human dataset are evaluated. Our dataset employs identical setup to ZJU-MoCap and additionally provides synchronized and calibrated multi-view videos with simple backgrounds and high-quality masks. The input image resolution is 512 × 512 for dynamic scenes.


\noindent
\textbf{Baselines.} For static scenes, comparisons are made with state-of-the-art generalizable radiance field methods \cite{25, 37, 38, 39} under both direct generalization and fine-tuning on particular scenes, including IBRNet \cite{38}, MVSNeRF \cite{39}, and ENeRF \cite{25}. The same settings as ENeRF are followed on the aforementioned benchmarks, with baseline results obtained from ENeRF. For dynamic scenes, comparisons are drawn against DNeRF \cite{67}, IBRNet, and ENeRF. For all baselines \cite{25, 38, 39, 67}, their released code is utilized with retraining under our experimental setting.


\subsection{Comparisons on view synthesis}
\noindent
\textbf{Comparisons on static scenes.} All baselines \cite{25, 38, 39} are compared on these datasets utilizing identical input views, with each scene tested on 4 novel views. Quantitative results for our model and other baselines on the NeRF Synthetic, Real Forward-facing, and DTU datasets are presented in Tab. \ref{t1}, demonstrating that our method achieves state-of-the-art performance. Qualitative results are shown in Fig. \ref{4} and \ref{5}. From the quantitative and qualitative results, it can be seen that our method achieves obvious performance optimization in better rendering effect than existing models. Specifically, improvements are attained at sampled boundaries by directly applying the pre-trained model and briefly fine-tuning. This signifies that rendering quality is enhanced by our ray sampling approach, especially in regions with enriched texture. Implementation of our ray sampling technique on other baselines \cite{38, 39} can also boost network convergence efficiency and rendering quality, as elaborated in the supplementary material.

\begin{figure}[t]
  \centering
  \includegraphics[width=1\linewidth]{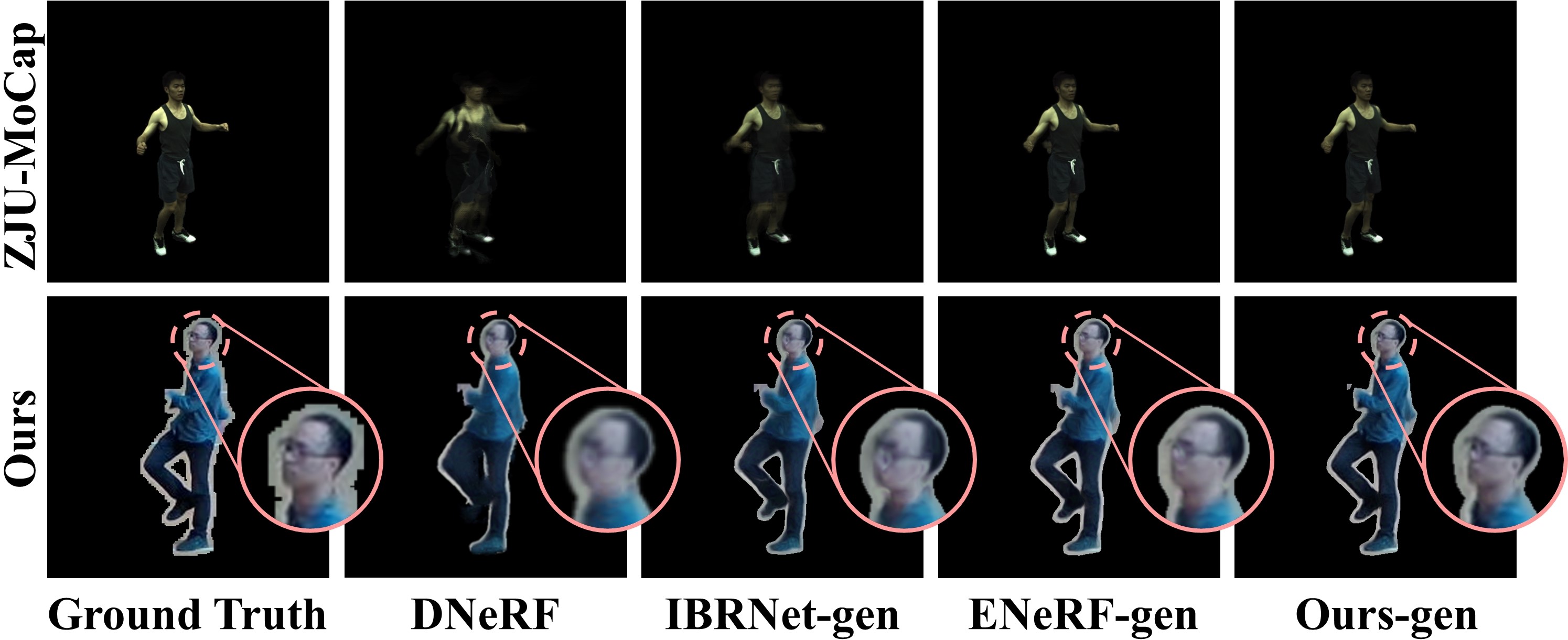}
   \caption{\textbf{Qualitative comparison of view synthesis results on dynamic scenes.} With the exception of DNeRF, all methods directly apply the pre-trained model to the scene without fine-tuning.}
   \label{6}
\end{figure}

\noindent
\textbf{Comparisons on dynamic scenes.} The ZJU-MoCap and our dataset provide multi-view human-body images with background removed. For DNeRF \cite{67}, identical settings to ENeRF are utilized, with the video sequence divided into several sub-sequences. We validate the performance of our method in modeling dynamic human bodies using ZJU-MoCap and our dataset. Quantitative results are summarized in Tab. \ref{t2}, while qualitative results for our proposed approach are depicted in Fig. \ref{6}. The quantitative analysis demonstrates our method realizes state-of-the-art performance after abbreviated fine-tuning time. This signifies our approach enhances training efficiency (less fine-tuning time) while sustaining significant rendering capability.

\begin{table}[t]
    \renewcommand{\arraystretch}{1}
    \centering
    \caption{\textbf{Quantitative results of dynamic scenes.} “0” indicates that the pre-trained model is directly used in the corresponding scene. All methods are evaluated on the same machine with corresponding fine-tuning time.}
    \label{t2}
    \scalebox{0.65}{
    \begin{tabular}{c|c|ccc|ccc}
    \Xhline{1.2pt}
        \multirow{2}{*}{Methods} & \multirow{2}{*}{\makecell{Training time}} & \multicolumn{3}{c|}{ZJU-MoCap} & \multicolumn{3}{c}{Our dataset} \\
        \cline{3-5} \cline{6-8} 
        ~ & ~ & PSNR$\uparrow$ & SSIM$\uparrow$ & LPIPS$\downarrow$ & PSNR$\uparrow$ & SSIM$\uparrow$ & LPIPS$\downarrow$ \\ 
        \hline
        DNeRF &10h&26.53&0.889&0.154&24.38&0.877&0.136\\
        IBRNet &0&29.46&0.947&0.094&26.89&0.920&0.065\\
        IBRNet &2h&32.38&0.968&0.065&28.03&0.926&\textbf{0.059}\\
        ENeRF &0&31.21&0.970&0.041&29.41&0.949&0.082\\
        ENeRF &2h&32.52&\textbf{0.978}&0.030&30.29&0.952&0.080\\
        Ours &0h&32.49&0.972&0.031&31.15&0.964&0.066\\
        Ours &1h&\textbf{33.61}&0.977&\textbf{0.025}&\textbf{31.81}&\textbf{0.969}&0.062\\
    \Xhline{1.2pt}
    \end{tabular}
    }
\end{table}

\subsection{Analysis of convergence efficiency}

\begin{figure}[t]
  \centering
  \includegraphics[width=1\linewidth]{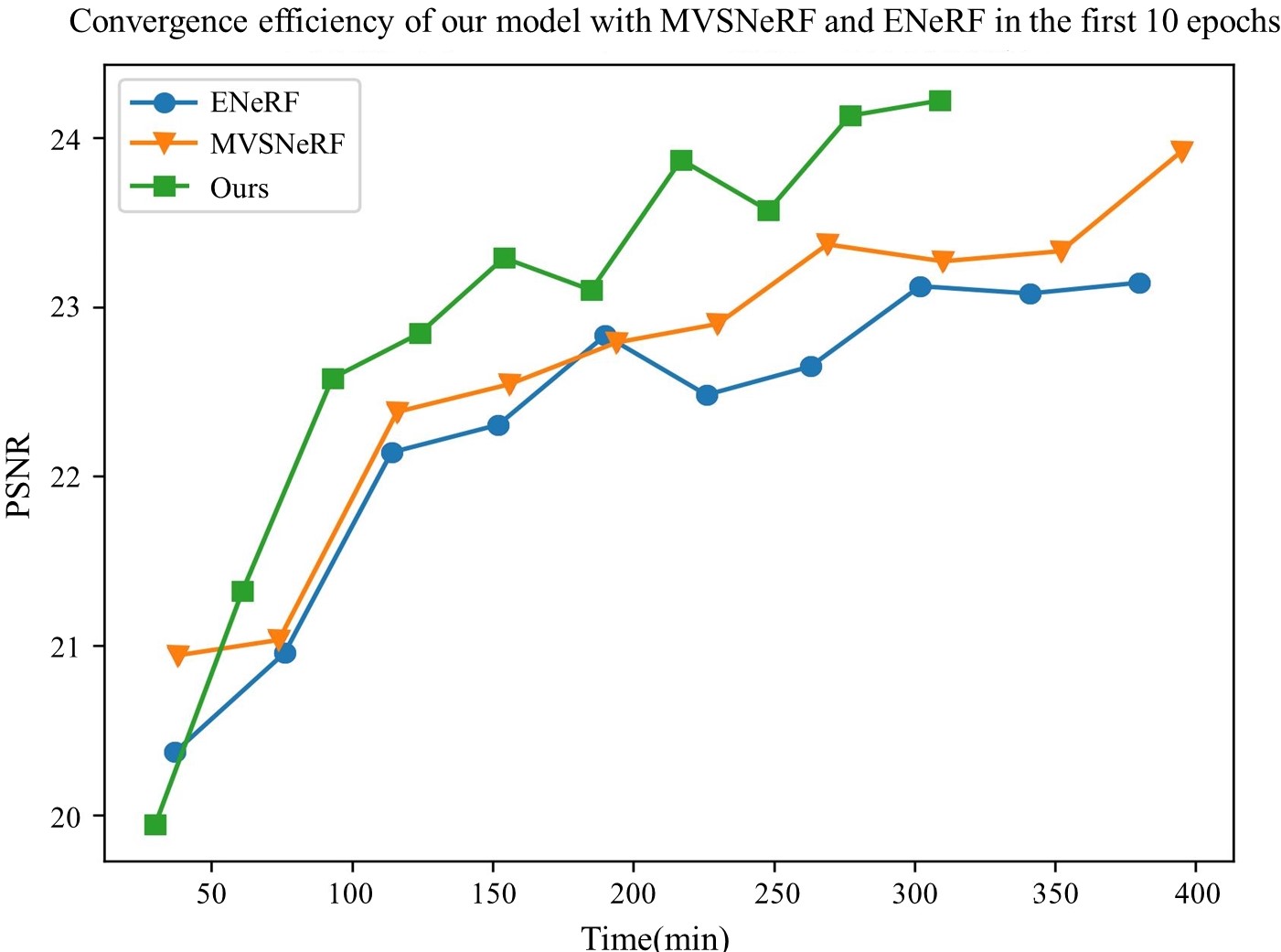}
   \caption{ \textbf{ Comparison of the efficiency of convergence of our method with other baselines.} Our model attains superior PSNR values (higher is preferable) given identical training durations, and accomplishes equivalent training results in markedly shorter convergence time versus MVSNeRF and ENeRF. This demonstrates that our approach achieves optimal convergence efficiency.}
   \label{7}
\end{figure}

Owing to the various ray sampling strategies proposed herein, the training efficiency of our model is markedly superior to existing generalizable neural radiance fields \cite{25,39,37,38}. The Adam optimizer \cite{63} is utilized for training on the RTX 3090 GPU with an initial learning rate of 5e-4, halving the learning rate after 30k iterations (1.7x faster than ENeRF). The model converges after 100k iterations (2x faster than ENeRF). Fig. \ref{7} presents a comparison of training time and PSNR for our proposed model, ENeRF, and MVSNeRF using the RTX 3090 GPU on the DTU dataset under random parameter initialization.

In this figure, merely the PSNR and time curves of the first 10 epochs are delineated. It can be discerned that the convergence rapidity of our proposed methodology is conspicuously superior to that of ENeRF and MVSNeRF during the training process. The predominant rationale is that we elect a more efficient ray sampling strategy, which allocates more training resources to regions with intricate information. This accelerates the convergence of the neural radiance fields on one hand. On the other hand, it also facilitates the radiance fields to perceive the scene depth more accurately, thereby ameliorating the final rendering quality.

\subsection{Ablation studies}

To compare the convergence acceleration afforded by the aforementioned ray sampling strategies, on the DTU dataset, we compare the convergence efficiency and results of ray sampling approaches guided by pixel regions and depth boundaries, respectively, against an adaptive ray sampling approach. The earliest convergence timepoints and associated rendering qualities for the different ray sampling strategies are appraised. Subsequently, the models are trained further without guided strategies until convergence, thereby evaluating the final performance enhancement conferred by the sampling strategies.

The first row of Tab. \ref{t3} delineates the convergence performance and time utilizing the default random ray sampling approach for neural radiance fields. The divergent sampling methodologies are ultimately trained for identical timespans to compare the final rendering qualities. As depicted in Tab. \ref{t3}, the proposed ray sampling strategies accelerate the convergence rate of neural radiance fields to varying extents. With adequate prolonged training, they conspicuously refine the rendering quality.

\begin{table}[t]
    \renewcommand{\arraystretch}{1}
    \centering
    \caption{\textbf{Quantitative ablation of the design choices on DTU dataset.} ``Pixel region" and ``Depth boundary" denote guiding the ray sampling from the neural radiance fields based on pixel regions and depth boundaries, respectively. The rendering resolution is configured to 512 × 640.}
    \label{t3}
    \scalebox{0.75}{
    \begin{tabular}{cccccc}
    \Xhline{1.2pt}
        \makecell{Pixel \\ region} & \makecell{Depth \\ boundary} & \makecell{Adaptive \\ ray sampling} & \makecell{PSNR\\(convergence)} &\makecell{Convergence \\ time(h)} & PSNR\\ \hline
        &&&27.53&22.5&27.53\\
        $\checkmark$&&&26.63&14.2&27.58\\
         &$\checkmark$&&26.97&13.8&27.61\\
         &&$\checkmark$&26.92&14.6&27.11\\
        $\checkmark$&$\checkmark$&&\textbf{27.65}&8.4&\textbf{27.85}\\
        $\checkmark$&$\checkmark$&$\checkmark$&27.43&\textbf{8.2}&27.55\\
    \Xhline{1.2pt}
    \end{tabular}
}
\end{table}

Among the approaches, the ray sampling methodology guided by pixel regions and depth boundaries demonstrates the most expeditious convergence, attaining comparable quality to the final rendering of the default random sampling upon convergence. These experimental findings corroborate our hypothesis regarding ray sampling redundancy, and validate the efficacy of our sampling strategy. The adaptive sampling approach analogously enhances training efficiency, akin to the online example mining technique in deep learning. However, the data indicates the convergence acceleration effect of adaptive sampling strategy is worse than our ray sampling strategies, with the combination of all three failing to confer additional improvements. This implies the sampling strategy guided by pixel regions and depth boundary subsumes the advantages of adaptive sampling, while proffering prior information more congruous with real-world scene observation, as reflected in the loss distribution. More details regarding the qualitative results of ablation experiments and adaptive ray sampling can be found in the supplementary material.

\section{Conclusion}

We propose a ray sampling approach that enhances both training efficiency and neural rendering quality. Our model samples rays under the guidance of pixel regions and depth boundaries, thereby densifying sampling in areas with greater color and depth variations. By concentrating training and fine-tuning on regions with intricate textures, our model effectively mitigates the redundancy of the sampled rays. The ENeRF framework \cite{25} integrating our ray sampling method demonstrates superior convergence efficiency and rendering quality compared to the vanilla ENeRF \cite{25}, while also generalizing better across diverse testing datasets with abbreviated training durations, surpassing concurrent works \cite{25,38,39,67}. With merely 2-3 input views, our model requires dramatically fewer iterations for fine-tuning on novel scenes to achieve competitive performance versus these baselines \cite{25,38,39,67}. Applying our sampling approach to existing models \cite{38,39} also confers improvements, further validating the broad utility of our method.


\clearpage
{\small
\bibliographystyle{ieee_fullname}
\bibliography{PaperForReview}
}

\clearpage
\appendix
\section*{Supplementary Material}
\section{Method Details}

\noindent
\textbf{Adaptive ray sampling.} The adaptive ray sampling approach enables the radiance fields to adaptively sample rays in regions with inaccurate rendering. This process is analogous to sample mining during training in conventional deep learning, striving to identify pixels with inadequate convergence via the loss function and confer greater priority.

The optimization of the adaptive ray sampling comprises two stages. Initially, a uniform distribution of pixels $s_i$ is randomly sampled from the input source views to update the network parameters and compute the loss. This can be achieved by partitioning each image into an [8 × 8] grid and calculating the mean loss within each square region $R_j$ $(j = 1,2, \cdots ,64)$:
\begin{equation}
    \mathcal{H}_i[j]=\frac{1}{|r_j|}\sum_{(u,v)\in r_j}{e_i^g[u,v]+e_i^p[u,v]},
\end{equation}
where $r_j=s_i\cap R_j$ represents the pixels uniformly sampled from $R_j$. These statistics are normalized into a probability distribution:
\begin{equation}
    f_i[j]=\frac{\mathcal{H}_i[j]}{\Sigma_{m=1}^{64}\mathcal{H}_i[m]}.
\end{equation}

\begin{figure}
    \centering  
    \includegraphics[width=1\linewidth]{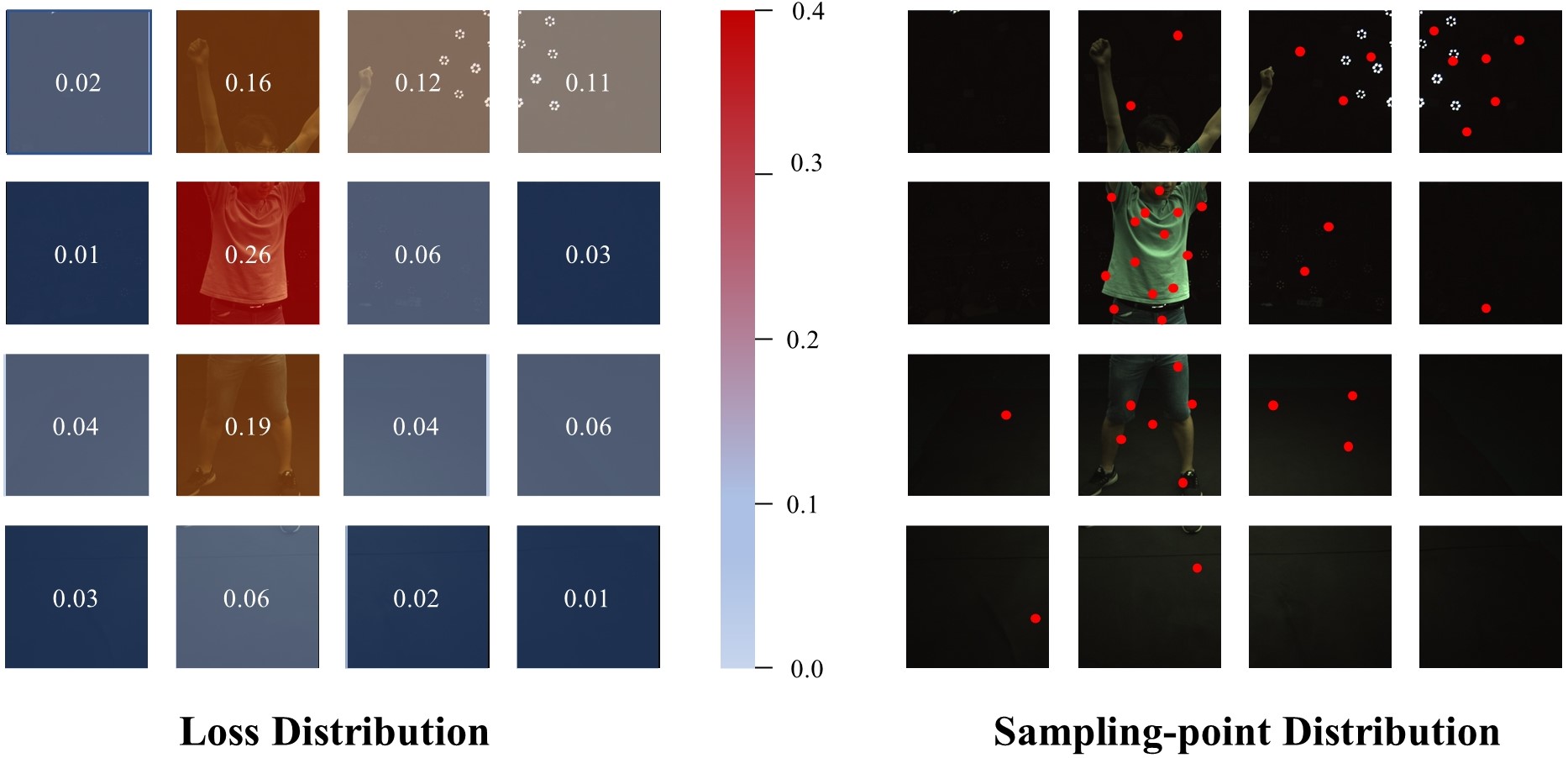}
    \caption{ \textbf{Sampling probability distribution of the adaptive ray sampling method.}}
    \label{9}
\end{figure}

This distribution is subsequently utilized to uniformly resample a number of pixels quantified by $n_i \cdot f_i[j]$ in each region (where $n_i$ represents the total sample count per image), thus allocating more samples to areas exhibiting elevated loss. The adaptive ray sampling approach is delineated in Fig. \ref{9}.

\begin{figure}
    \centering
    \includegraphics[width=\linewidth]{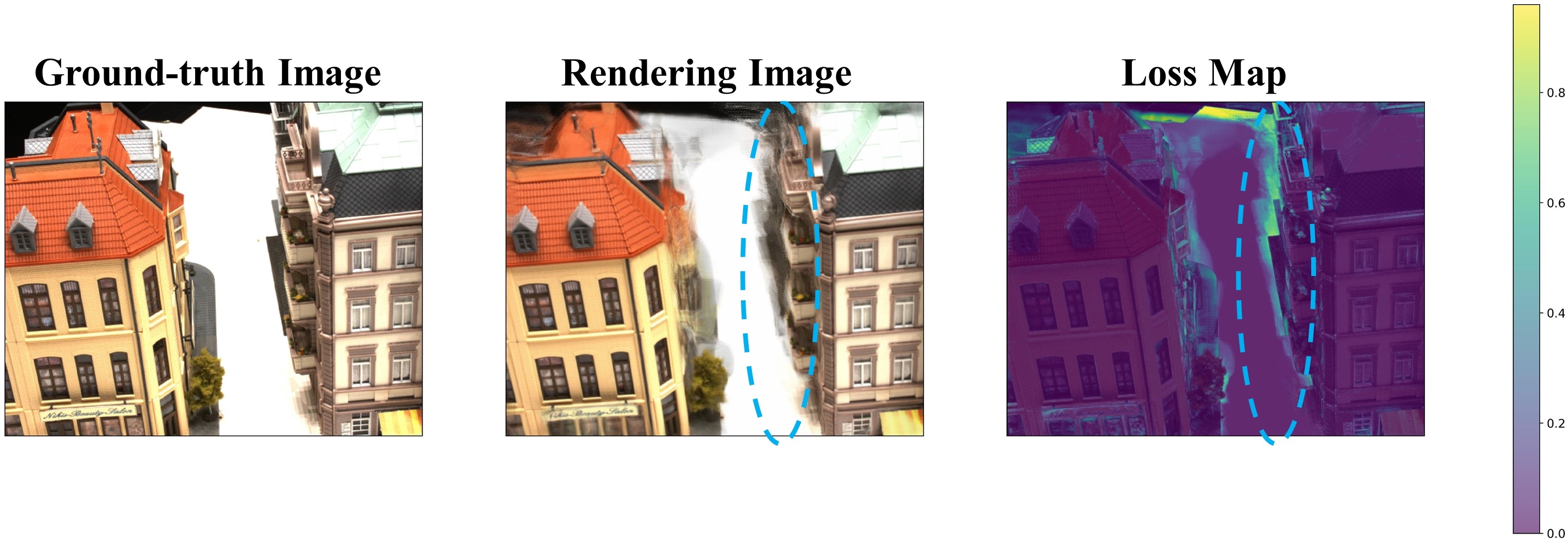}
    \caption{ \textbf{The training process of NeRF.} Regions with sharp texture variation exhibit larger loss values in loss map. Correspondingly, these areas demonstrate poorer rendering performance in rendering image. Experiments reveal areas of more detailed texture require prolonged training time, with slower convergence rate.}
    \label{2}
\end{figure}

\noindent
\textbf{Analysis of training convergence.} The loss distribution of NeRF model during the training process is analyzed, and it is found that the fitting efficiency is slower in the image region with drastic color changes than in other regions. This difference shows significant detail texture and depth changes in the 3D scene area. As shown in Fig. \ref{2}, this phenomenon aligns with rendering characteristics of NeRF.

\begin{figure}
    \centering  
    \includegraphics[width=1\linewidth]{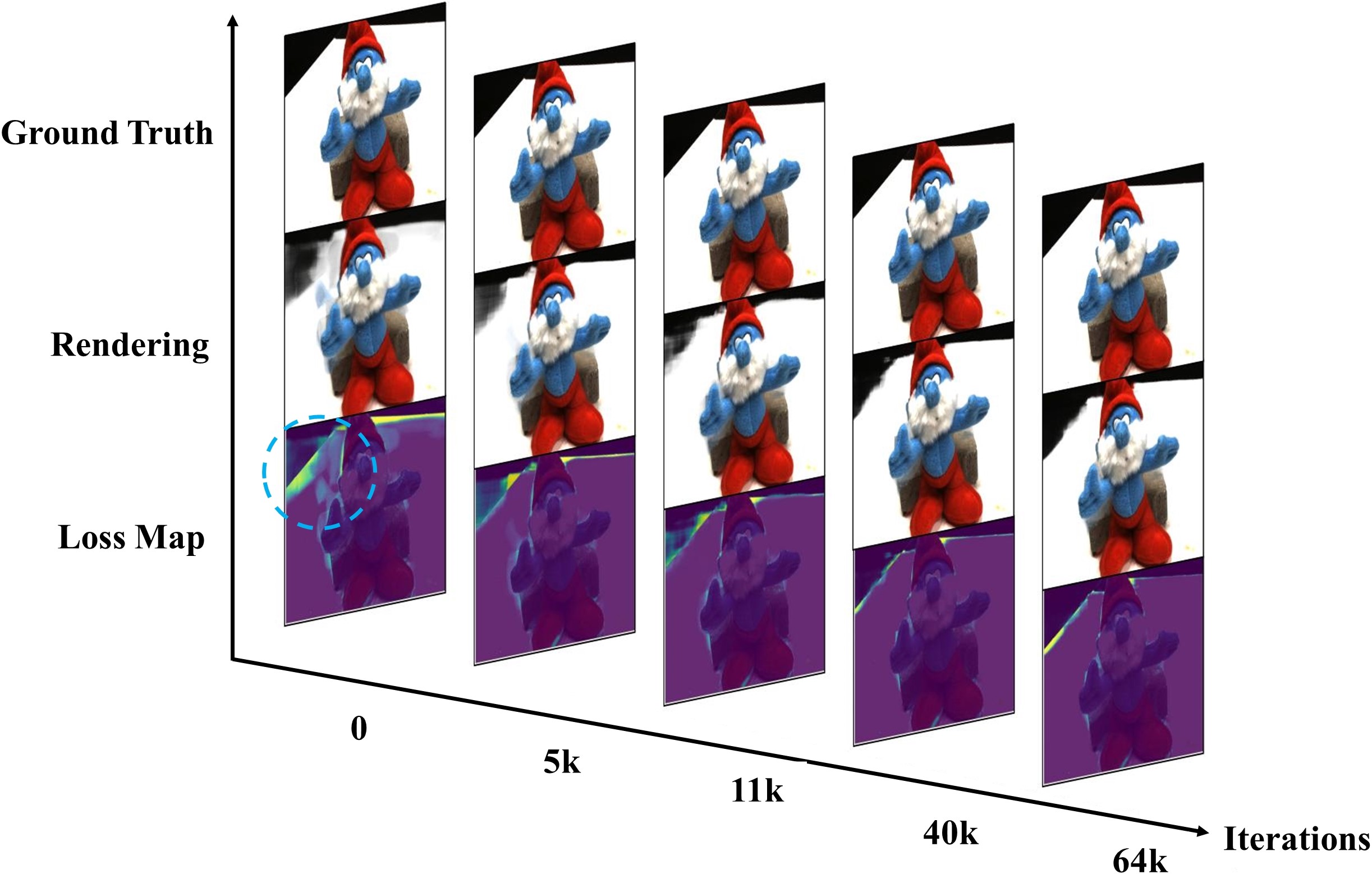}
    \caption{ \textbf{Training process and loss distribution.} It is evident that regions containing simple details exhibit rapid convergence within 64 epochs of training. However, complex areas maintain substantial loss values necessitating further convergence, yielding relatively inferior renderings.}
    \label{10}
\end{figure}

Fig. \ref{10} demonstrates the learning rate of the radiance fields converges more quickly in regions exhibiting negligible color and depth variations. However, areas with pronounced color and depth variations, typifying complex scenes, necessitate extended durations for detailed rendering.

This analysis is further substantiated by examining the loss distribution during ENeRF \cite{25} training. As depicted in the third row of Fig. \ref{10}, regions with sparse textures predominantly converge early in the training phase, while greater time is requisite for gradual convergence in areas with dense depth boundaries and color textures. This indicates redundancy in the existing ray sampling strategy, directly hampering the convergence efficiency of radiance fields.

\begin{figure}
    \centering  
    \includegraphics[width=1\linewidth]{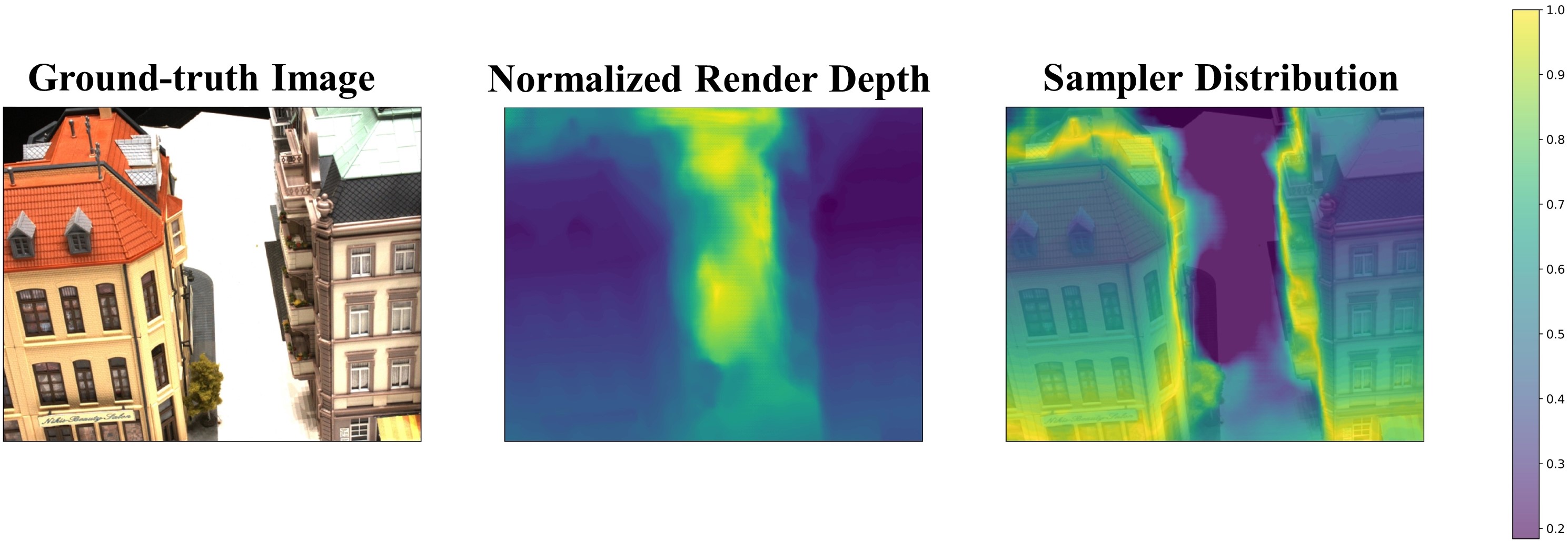}
    \caption{ \textbf{The distribution of depth-guided ray sampling method.} }
    \label{11}
\end{figure}

\begin{figure}
    \centering  
    \includegraphics[width=1\linewidth]{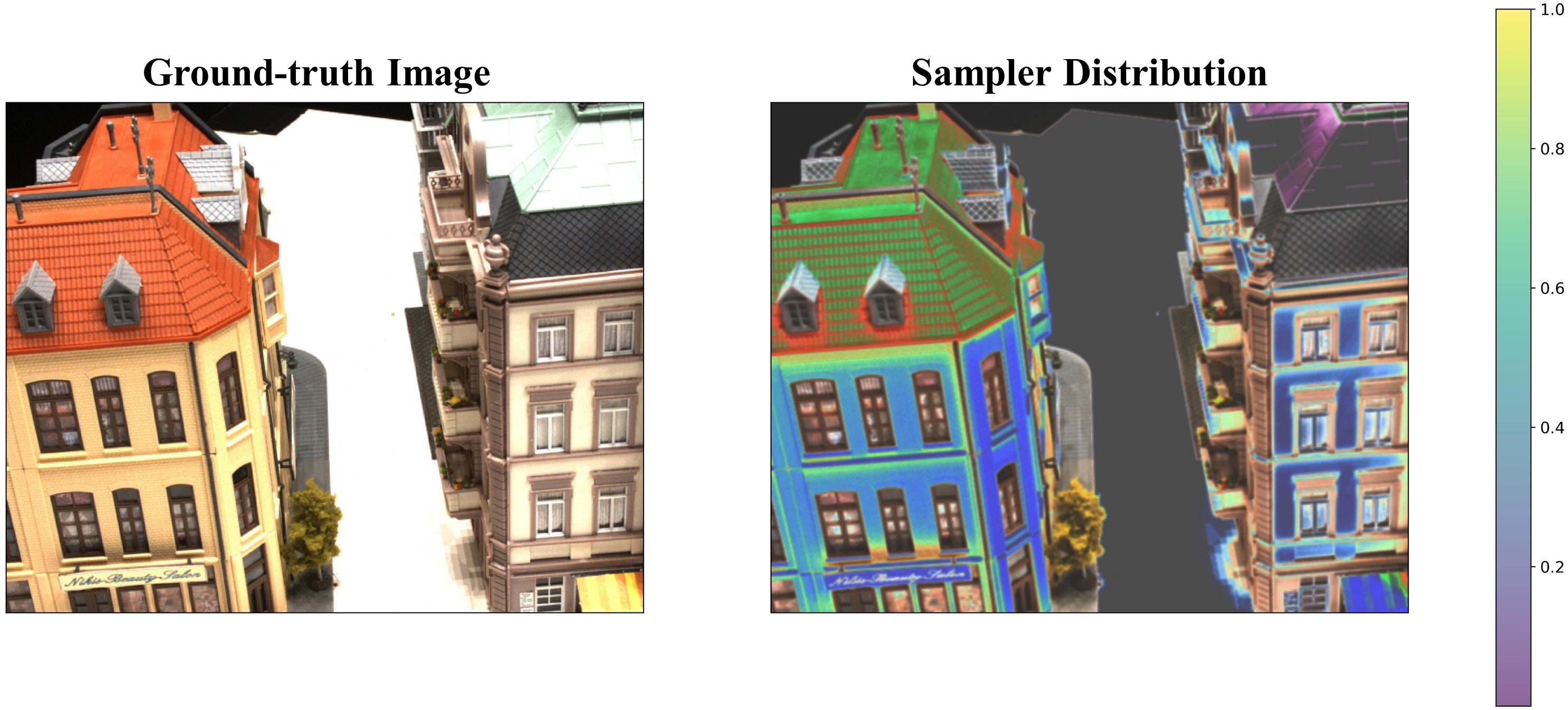}
    \caption{ \textbf{The distribution of pixel-guided ray sampling method.} }
    \label{12}
\end{figure}

\noindent
\textbf{Our ray sampling methods.} Employing depth-guided ray sampling, we sample more rays in regions exhibiting pronounced depth variations. As portrayed in Fig. \ref{11}, greater sampling probability is conferred at depth boundaries, where heightened depth fluctuations with richer details necessitate densified ray sampling during the training stage. We additionally implement pixel-guided sampling, assigning higher probabilities when the color difference between the pixel $(u, v)$ and its neighbors is substantial, as demonstrated in Fig. \ref{12}. This vividly exemplifies the sampling methodology under our proposed pixel-guided approach.

\noindent
\textbf{Clamp function.}  The $clamp$ function will return the parameter value if the parameter is within the numerical range between the minimum and maximum values, the maximum value if the parameter is larger than the range, and the minimum value if the parameter is smaller than the range. Such as $clamp(minnumber, maxnumber, parameter)$. $clamp(4,6,22)$ returns 6 because 22 is greater than 6 and 6 is the largest value in the range, $clamp(4,6,2)$ returns 4 because 2 is less than 4 and 4 is the smallest value in the range, $clamp(4,6,5)$ returns 5 because the value is in the range. 


\section{More Experimental Results}

\begin{figure}[t]
  \centering
  \includegraphics[width=1\linewidth]{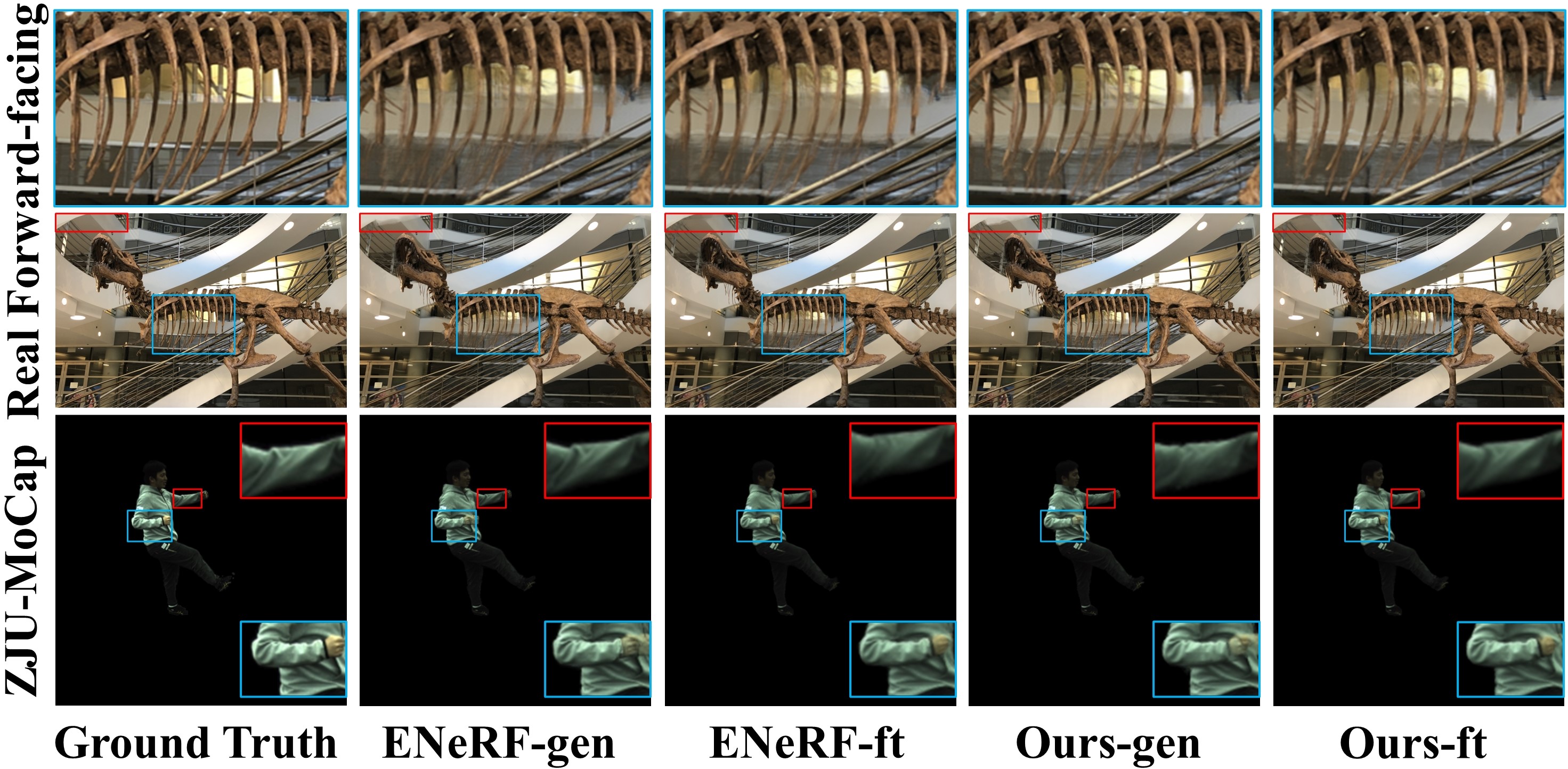}
   \caption{ \textbf{Qualitative comparison highlighting improvements in regions with abundant textures.} ``-gen" denotes that these pre-trained models are directly deployed on the input images without fine-tuning. ``-ft" signifies that these pre-trained models undergo fine-tuning on the provided scenes for 2k iterations.}
   \label{8}
\end{figure}

\noindent
\textbf{Analysis of rendering quality.} Our ray sampling approach samples more rays in regions with intricate details, as reflected in dramatic variations of color and depth. In this manner, the scene geometry is perceived with enhanced accuracy, and greater attention is conferred to areas with rich texture, consequently engendering marked improvements compared to the vanilla ENeRF model devoid of our method, as exemplified in Fig. \ref{8}. By comparing the generalization and fine-tuning results between our model and ENeRF \cite{25}, with all models fine-tuned for 2k iterations, our model demonstrates superior details compared to ENeRF. Therefore, our model not only expedites the network convergence, but also augments rendering quality under analogous convergence circumstances.

\begin{figure*}
    \centering  
    \includegraphics[width=1\linewidth]{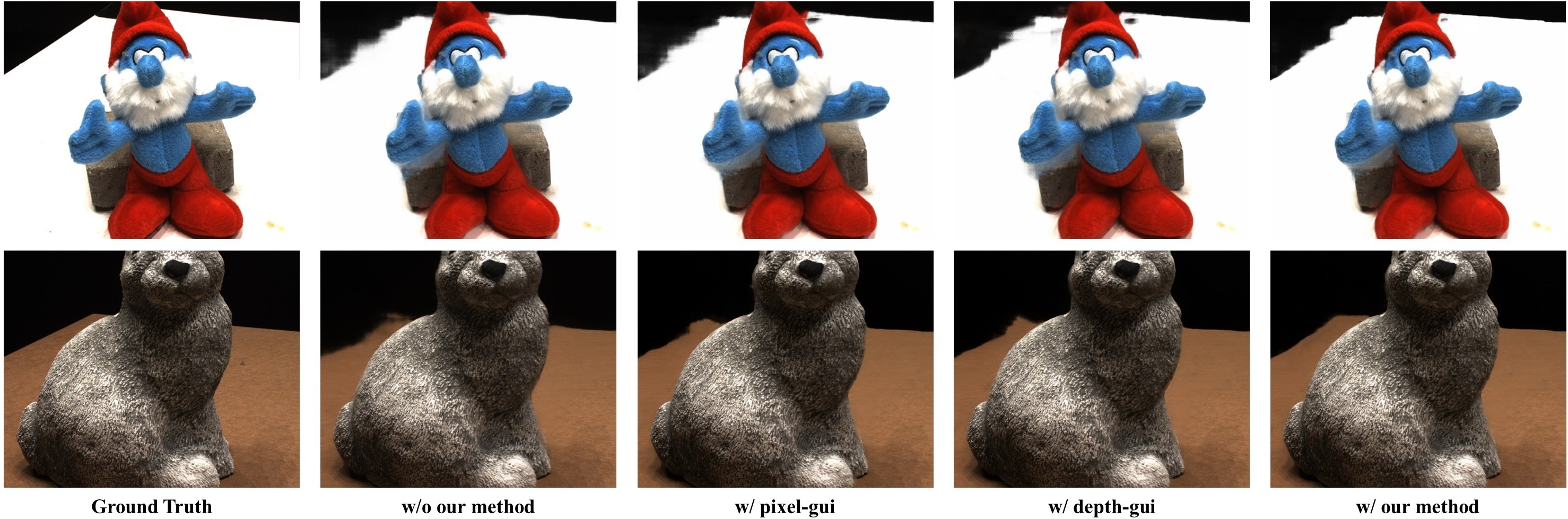}
    \caption{ \textbf{Visual ablation results on DTU dataset. ``w/o our method'' is similar to ENeRF \cite{25}}. }
    \label{14}
\end{figure*}

\noindent
\textbf{Visual ablation results.} Visual ablation results are presented in Fig. \ref{14}. Relative to uniform sampling strategy, superior rendering is achievable solely utilizing either pixel-guided or depth-guided approaches, at accelerated training velocities. The combination of both ray sampling methodologies, with ray sampling under the joint guidance of pixel region and depth boundary, confers additional refinements in rendering quality and convergence efficiency compared to uncombined ray sampling methodology, as portrayed in Fig. \ref{14}.

\begin{table}[t]
    \renewcommand{\arraystretch}{1.3}
    \centering
    \caption{\textbf{Quantitative results with these models on the NeRF Synthesis and DTU datasets.} The means of ``ori'' and ``ft'' are the same as Fig. \ref{13}}
    \label{t4}
    \scalebox{0.55}{
    \begin{tabular}{c|c|ccc|ccc}
    \Xhline{1.2pt}
        \multirow{2}{*}{Methods} & \multirow{2}{*}{\makecell{Training settings}} & \multicolumn{3}{c|}{DTU} & \multicolumn{3}{c}{NeRF Synthetic} \\
        \cline{3-5} \cline{6-8}
        ~ & ~ & PSNR$\uparrow$ & SSIM$\uparrow$ & LPIPS$\downarrow$ & PSNR$\uparrow$ & SSIM$\uparrow$ & LPIPS$\downarrow$ \\ 
        \hline
        IBRNet-ori & \multirow{4}{*}{\makecell{Generalization \\ (only pre-trained)}} &26.04&0.917&0.191&22.44&0.874&0.195\\
        MVSNeRF-ori & ~ &26.63&0.931&0.168&23.62&0.897&0.176 \\ 
        IBRNet-ft & ~ &27.38&0.937&0.166&23.77&0.915&0.181\\
        MVSNeRF-ft & ~ &\textbf{27.53}&\textbf{0.940}&\textbf{0.143}&\textbf{24.83}&\textbf{0.924}&\textbf{0.172}\\
        \hline
        IBRNet-ori(1h) & \multirow{4}{*}{\makecell{Fine-tuning \\ on given scene}} &31.35&0.956&0.131&25.62&0.939&0.111 \\
        MVSNeRF-ori(15min) & ~ &28.51&0.933&0.179&27.07&0.931&0.168 \\ 
        IBRNet-ft(1h) & ~ &\textbf{31.90}&\textbf{0.958}&\textbf{0.122}&26.82&\textbf{0.944}&\textbf{0.102}\\
        MVSNeRF-ft(15min) & ~ &29.28&0.941&0.167&\textbf{28.10}&0.936&0.155\\
    \Xhline{1.2pt}
    \end{tabular}
    }
\end{table}

\noindent
\textbf{Other baselines with our method.} Ray sampling constitutes a ubiquitous process within NeRF modeling paradigms; therefore, our proposed ray sampling method can be directly integrated into various extant NeRF frameworks \cite{37, 38, 39} to enhance training efficiency and rendering fidelity. To demonstrate the generalization capacity, we incorporate our approach into MVSNeRF \cite{39} and IBRNet \cite{38} separately and evaluate the modified baselines on NeRF Synthesis and DTU datasets relative to the original methods. Experiments adopt identical configurations as described previously.

\begin{figure*}
    \centering  
    \includegraphics[width=0.99\linewidth]{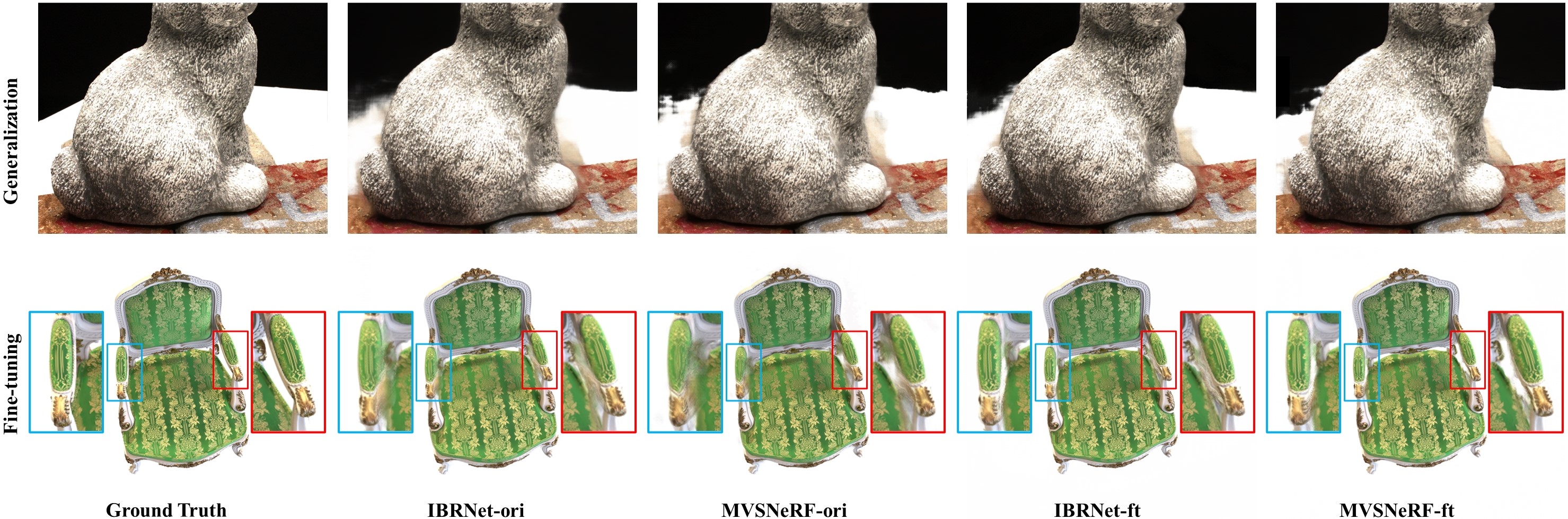}
    \caption{ \textbf{Qualitative results on NeRF Synthesis and DTU datasets.} “IBRNet-ori” and “MVSNeRF-ori” mean that these models without our ray sampling method. “IBRNet-ft” and “MVSNeRF-ft” mean that these models are modified with our method.}
    \label{13}
\end{figure*}

Integrating our ray sampling methodology engenders 1.7× (IBRNet) and 2× (MVSNeRF) accelerations in convergence speed versus the original frameworks. Guidance from pixel regions and depth boundaries in sampling confers additional gains in training speed alongside superior rendering fidelity. By concentrating sampling probabilities in areas exhibiting pronounced depth and color variations given equal fine-tuning durations, qualitative outcomes after generalization and fine-tuning showcase marked improvements in these regions, as portrayed in Fig. \ref{13}. Quantitative results are tabulated in Tab. \ref{t4}.

\begin{table}[t]
    \centering
    \caption{\textbf{Quantitative comparison on the DTU dataset.} }
    \label{t5}
    \scalebox{0.9}{
    \begin{tabular}{c|ccccc}
    \Xhline{1.2pt}
        Scan & $\#$1 \ & $\#$8 & $\#$21 & $\#$103 & $\#$114 \\
        \hline
         Metric & \multicolumn{5}{c}{PSNR$\uparrow$} \\ 
        \hline
        IBRNet &25.97 &27.45 &20.94 &27.91 &27.91 \\
        MVSNeRF &26.96 &27.43 &21.55 &29.25 &27.99 \\ 
        ENeRF &28.85 &29.05 &22.53 &30.51 &28.86 \\
        Ours &\textbf{29.43} &\textbf{29.62} &\textbf{24.35} &\textbf{30.79} &\textbf{28.92} \\
    \hline
        IBRNet-ft(1h) &31.00 &\textbf{32.46} &\textbf{27.88} &\textbf{34.40} &\textbf{31.00} \\
        MVSNeRF-ft(15min) &28.05 &28.88 &24.87 &32.23 &28.47 \\ 
        ENeRF-ft(15min) &29.81 &30.06 &22.50 &31.57 &29.72 \\
        ENeRF-ft(1h)  &30.10 &30.50 &22.46 &31.42 &29.87 \\
        Ours-ft(15min) &30.87 &30.32 &24.89 &32.01 &30.28 \\
        Ours-ft(30min)  &\textbf{31.11} &30.88 &24.95 &32.25 &30.97 \\
    \hline
         Metric & \multicolumn{5}{c}{SSIM$\uparrow$} \\ 
        \hline
        IBRNet &0.918 &0.903 &0.873 &0.950 &0.943 \\
        MVSNeRF &0.937 &0.922 &0.890 &0.962 &0.949 \\ 
        ENeRF &0.958 &0.955 &0.916 &0.968 &\textbf{0.961} \\
        Ours &\textbf{0.963} &\textbf{0.956} &\textbf{0.931} &\textbf{0.969} &0.960 \\
    \hline
        IBRNet-ft(1h) &0.955 &0.945 &\textbf{0.947} &0.968 &0.964 \\
        MVSNeRF-ft(15min) &0.934 &0.900 &0.922 &0.964 &0.945 \\ 
        ENeRF-ft(15min) &0.964 &0.958 &0.922 &0.971 &0.965 \\
        ENeRF-ft(1h)  &0.966 &0.959 &0.924 &0.971 &\textbf{0.965} \\
        Ours-ft(15min) &0.967 &0.964 &0.935 &0.977 &0.962 \\
        Ours-ft(30min)  &\textbf{0.968} &\textbf{0.967} &0.935 &\textbf{0.980} &0.964 \\
    \hline
        Metric & \multicolumn{5}{c}{LPIPS$\downarrow$} \\ 
        \hline
        IBRNet &0.190 &0.252 &0.179 &0.195 &0.136 \\
        MVSNeRF &0.155 &0.220 &0.166 &0.165 &0.135 \\ 
        ENeRF &0.086 &0.119 &0.107 &0.107 &0.076 \\
        Ours &\textbf{0.078} &\textbf{0.114} &\textbf{0.089} &\textbf{0.104} &\textbf{0.075} \\
    \hline
        IBRNet-ft(1h) &0.129 &0.170 &0.104 &0.156 &0.099 \\
        MVSNeRF-ft(15min) &0.171 &0.261 &0.142 &0.170 &0.153 \\ 
        ENeRF-ft(15min) &0.074 &0.109 &0.100 &0.103 &0.075 \\
        ENeRF-ft(1h)  &0.071 &0.106 &0.097 &0.102 &0.074 \\
        Ours-ft(15min) &0.059 &0.107 &0.074 &0.100 &0.072 \\
        Ours-ft(30min)  &\textbf{0.57} &\textbf{0.105} &\textbf{0.71} &\textbf{0.098} &\textbf{0.072} \\
    \Xhline{1.2pt}
    \end{tabular}
    }
\end{table}

\section{Per-scene breakdown}

Tab. \ref{t5}, \ref{t6}, \ref{t7} delineate the per-scene breakdown of the quantitative results presented in the manuscript for the NeRF Synthetic, DTU and Real Forward-facing datasets, respectively. These results align with the averaged metrics depicted in the paper. The quantitative data demonstrates our methodology attaining comparable performance to other baselines. The baseline results are obtained from ENeRF \cite{25}. 

\begin{table*}[t]
    \centering
    \caption{\textbf{Quantitative comparison on the NeRF Synthetic dataset.} }
    \label{t6}
    \scalebox{1}{
    \begin{tabular}{c|cccccccc}
    \Xhline{1.2pt}
        &Chair &Drums &Ficus &Hotdog &Lego &Materials &Mic &Ship \\
        \hline
         Metric & \multicolumn{8}{c}{PSNR$\uparrow$} \\ 
        \hline
        IBRNet &24.20 &18.63 &21.59 &27.70 &22.01 &20.91 &22.10 &22.36 \\
        MVSNeRF &23.35 &20.71 &21.98 &28.44 &23.18 &20.05 &22.62 &23.35 \\ 
        ENeRF &28.44 &24.55 &23.86 &34.64 &24.98 &24.04 &26.60 &26.09 \\
        Ours &\textbf{29.43} &\textbf{25.77} &\textbf{25.12} &\textbf{34.82} &\textbf{25.25} &\textbf{24.41} &\textbf{27.00} &\textbf{26.32}\\
    \hline
        IBRNet-ft(1h) &28.18 &21.93 &25.01 &31.48 &25.34 &24.27 &27.29 &21.48 \\
        MVSNeRF-ft(15min) &26.80 &22.48 &26.24 &32.65 &26.62 &25.28 &29.78 &26.73 \\ 
        ENeRF-ft(15min) &28.52 &25.09 &24.38 &35.26 &25.31 &24.91 &28.17 &25.95 \\
        ENeRF-ft(1h)  &28.94 &25.33 &24.71 &\textbf{35.63} &25.39 &24.98 &29.25 &26.36 \\
        Ours-ft(15min) &29.61 &26.32 &26.44 &35.44 &25.97 &25.38 &29.96 &26.73 \\
        Ours-ft(30min)  &\textbf{30.12} &\textbf{26.47} &\textbf{26.85} &35.56 &\textbf{26.08} &\textbf{25.51} &\textbf{30.01} &\textbf{27.05} \\
    \hline
         Metric & \multicolumn{8}{c}{SSIM$\uparrow$} \\ 
        \hline
        IBRNet &0.888 &0.836 &0.881 &0.923 &0.874 &0.872 &0.927 &0.794 \\
        MVSNeRF &0.876 &0.886 &0.898 &0.962 &0.902 &0.893 &0.923 &0.886 \\ 
        ENeRF &0.966 &0.953 &0.931 &0.982 &0.949 &0.937 &\textbf{0.971} &0.893 \\
        Ours &\textbf{0.968} &\textbf{0.956} &\textbf{0.934} &\textbf{0.983} &\textbf{0.952} &\textbf{0.938} & 0.969 &\textbf{0.896} \\
    \hline
        IBRNet-ft(1h) &0.955 &0.913 &0.940 &0.978 &0.940 &0.937 &0.974 &0.877 \\
        MVSNeRF-ft(15min) &0.934 &0.898 &0.944 &0.971 &0.924 &0.927 &0.970 &0.879 \\ 
        ENeRF-ft(15min) &0.968 &0.958 &0.936 &0.984 &0.948 &0.946 &0.981 &0.891 \\
        ENeRF-ft(1h)  &0.971 &\textbf{0.960} &\textbf{0.939} &\textbf{0.985} &0.949 &\textbf{0.947} &\textbf{0.985} &0.893 \\
        Ours-ft(15min) &0.964 &0.955 &0.932 &0.970 &0.940 &0.934 &0.972 &0.885 \\
        Ours-ft(30min)  &\textbf{0.972} &0.959 &0.938 &0.984 &\textbf{0.949} &0.945 &0.983 &\textbf{0.894} \\
    \hline
        Metric & \multicolumn{8}{c}{LPIPS$\downarrow$} \\ 
        \hline
        IBRNet &0.144 &0.241 &0.159 &0.175 &0.202 &0.164 &0.103 &0.369 \\
        MVSNeRF &0.282 &0.187 &0.211 &0.173 &0.204 &0.216 &0.177 &0.244 \\ 
        ENeRF &0.043 &0.056 &0.072 &0.039 &0.075 &0.073 &0.040 &0.181 \\
        Ours &\textbf{0.032} &\textbf{0.044} &\textbf{0.065} &\textbf{0.039} &\textbf{0.073} &\textbf{0.065} &\textbf{0.039} &\textbf{0.167} \\
    \hline
        IBRNet-ft(1h) &0.079 &0.133 &0.082 &0.093 &0.105 &0.093 &0.040 &0.257 \\
        MVSNeRF-ft(15min) &0.129 &0.197 &0.171 &0.094 &0.176 &0.167 &0.117 &0.294 \\ 
        ENeRF-ft(15min) &0.033 &0.047 &0.069 &0.031 &0.073 &0.063 &0.021 &0.190 \\
        ENeRF-ft(1h)  &0.030 &0.045 &0.071 &0.028 &0.070 &0.059 &0.017 &0.183\\
        Ours-ft(15min) &0.032 &0.046 &0.068 &0.032 &0.070 &0.065 &0.020 &0.188 \\
        Ours-ft(30min)  &\textbf{0.028} &\textbf{0.041} &\textbf{0.070} &\textbf{0.027} &\textbf{0.067} &\textbf{0.056} &\textbf{0.015} &\textbf{0.162} \\
    \Xhline{1.2pt}
    \end{tabular}
    }
\end{table*}

\clearpage

\begin{table*}[t]
    \centering
    \caption{\textbf{Quantitative comparison on the Real Forward-facing dataset.} }
    \label{t7}
    \scalebox{1}{
    \begin{tabular}{c|cccccccc}
    \Xhline{1.2pt}
        &Fern &Flower &Fortress &Horns &Leaves &Orchids &Room &Trex \\
        \hline
         Metric & \multicolumn{8}{c}{PSNR$\uparrow$} \\ 
        \hline
        IBRNet &20.83 &22.38 &27.67 &22.06 &18.75 &15.29 &27.26 &20.06 \\
        MVSNeRF &21.15 &24.74 &26.03 &23.57 &17.51 &17.85 &26.95 &23.20 \\ 
        ENeRF &20.88 &24.78 &28.63 &23.51 &17.78 &17.34 &28.94 &20.37 \\
        Ours &\textbf{21.14} &\textbf{25.85} &\textbf{28.70} &\textbf{23.53} &\textbf{18.02} &\textbf{17.41} &\textbf{28.94} &\textbf{20.62}\\
    \hline
        IBRNet-ft(1h) &22.64 &26.55 &30.34 &25.01 &22.07 &19.01 &31.05 &22.34 \\
        MVSNeRF-ft(15min) &23.10 &27.23 &30.43 &26.35 &21.54 &20.51 &30.12 &24.32 \\ 
        ENeRF-ft(15min) &21.84 &27.46 &29.58 &24.97 &20.95 &19.17 &29.73 &23.01 \\
        ENeRF-ft(1h)  &22.08 &27.74 &29.58 &25.50 &21.26 &19.50 &30.07 &23.39 \\
        Ours-ft(15min) &22.53 &27.90 &29.54 &25.48 &21.67 &19.76 &30.04 &23.55 \\
        Ours-ft(30min)  &\textbf{23.42} &\textbf{28.37} &\textbf{30.88} &\textbf{26.17} &\textbf{22.84} &\textbf{21.02} &\textbf{31.01} &\textbf{24.94} \\
    \hline
         Metric & \multicolumn{8}{c}{SSIM$\uparrow$} \\ 
        \hline
        IBRNet &0.710 &0.854 &0.894 &0.840 &0.705 &0.571 &0.950 &0.768 \\
        MVSNeRF &0.638 &0.888 &0.872 &0.868 &0.667 &0.657 &0.951 &0.868 \\ 
        ENeRF &0.727 &0.890 &0.920 &\textbf{0.866} &0.685 &\textbf{0.637} &0.958 &0.778 \\
        Ours &\textbf{0.742} &\textbf{0.904} &\textbf{0.931} &0.864 &\textbf{0.692} &0.635 &\textbf{0.962} &\textbf{0.782} \\
    \hline
        IBRNet-ft(1h) &0.774 &0.909 &0.937 &0.904 &0.843 &0.705 &0.972 &0.842 \\
        MVSNeRF-ft(15min) &0.795 &0.912 &0.943 &0.917 &0.826 &0.732 &0.966 &0.895 \\ 
        ENeRF-ft(15min) &0.758 &0.919 &0.940 &0.893 &0.816 &0.710 &0.963 &0.855 \\
        ENeRF-ft(1h)  &0.770 &0.923 &\textbf{0.940} &0.904 &0.827 &0.725 &0.965 &0.869 \\
        Ours-ft(15min) &0.764 &0.920 &0.938 &0.898 &0.821 &0.717 &0.961 &0.863 \\
        Ours-ft(30min)  &\textbf{0.791} &\textbf{0.944} &0.938 &\textbf{0.925} &\textbf{0.843} &\textbf{0.750} &\textbf{0.982} &\textbf{0.874} \\
    \hline
        Metric & \multicolumn{8}{c}{LPIPS$\downarrow$} \\ 
        \hline
        IBRNet &0.349 &0.224 &0.196 &0.285 &0.292 &0.413 &0.161 &0.314 \\
        MVSNeRF &0.238 &0.196 &0.208 &0.237 &0.313 &0.274 &0.172 &0.184 \\ 
        ENeRF &0.235 &0.168 &0.118 &0.200 &0.245 &0.308 &0.141 &0.259 \\
        Ours &\textbf{0.222} &\textbf{0.153} &\textbf{0.107} &\textbf{0.185} &\textbf{0.232} &\textbf{0.293} &\textbf{0.140} &\textbf{0.248} \\
    \hline
        IBRNet-ft(1h) &0.266 &0.146 &0.133 &0.190 &0.180 &0.286 &0.089 &0.222 \\
        MVSNeRF-ft(15min) &0.253 &0.143 &0.134 &0.188 &0.222 &0.258 &0.149 &0.187 \\ 
        ENeRF-ft(15min) &0.220 &0.130 &0.103 &0.177 &0.181 &0.266 &0.123 &0.183 \\
        ENeRF-ft(1h)  &0.197 &0.121 &0.101 &0.155 &0.168 &0.247 &0.113 &0.169\\
        Ours-ft(15min) &0.206 &0.129 &0.114 &0.158 &0.180 &0.254 &0.121 &0.175 \\
        Ours-ft(30min)  &\textbf{0.185} &\textbf{0.106} &\textbf{0.081} &\textbf{0.142} &\textbf{0.157} &\textbf{0.238} &\textbf{0.113} &\textbf{0.158} \\
    \Xhline{1.2pt}
    \end{tabular}
    }
\end{table*}

\end{document}